# A BIOMIMETIC APPROACH BASED ON IMMUNE SYSTEMS FOR CLASSIFICATION OF UNSTRUCTURED DATA

Reda Mohamed HAMOU    Abdelmalek AMINE    Ahmed Chaouki LOKBANI

*Matematics and Computer Science Department*
*Tahar MOULAY University of Saïda, Algeria*

**Abstract.** In this paper we present the results of unstructured data clustering in this case a textual data from Reuters 21578 corpus with a new biomimetic approach using immune system. Before experimenting our immune system, we digitalized textual data by the n-grams approach. The novelty lies on hybridization of n-grams and immune systems for clustering. The experimental results show that the recommended ideas are promising and prove that this method can solve the text clustering problem.

## 1. Introduction

The Text mining is the combination of techniques and methods for the automatic processing of textual data in natural language. It is a multidimensional analysis of textual data, which aims to analyze and discover knowledge and connections from the available documents. In text mining similarities are used to produce synthetic representations of large collection of documents. Text mining includes a series of steps to go documents to text, text to number, number to the analysis and analysis to decision-making.

The problem of classification of data is identified as one of the major problems in extracting knowledge from data.

The popularity and complexity of data classification problem gave birth to a multitude of methods of resolution. These methods can both make use of heuristics or mathematical principles. Among them there is a branch that is inspired by biology, called bio-inspired methods.

The aim of our study is to clustering of text documents from the Reuters corpus by immune systems. The context was to prove that this biomimetic tool was capable of effectively regroups a similar documents. Obtaining acceptable results in our opinion we increase the number of documents to classify and the results were satisfactory. Our work has been validated by one evaluation function in this case the F-measure plus a comparison was made with a method derived from artificial intelligence is the cellular automata.

Biomimetic in a literary sense is the imitation of life. Biology has always been a source of inspiration for researchers in different fields. These have found an almost ideal in the observation of natural phenomena and their adaptation to solve problems. Among these models are the genetic algorithms, ant colonies, and swarms particles, clouds of flying insects [13] and of course immune systems that we will detail in the next section. The first approaches mentioned methods are widely recognized and studied but immune systems against methods are rarely used and in particular in the field of clustering. It has been our motivation for the use of this method in this field.

In extracting knowledge from data, the problem of data classification (clustering) is regarded as one of its main problematic.

Section 1 gives an introduction and state of the art, Section 2 presents representation of texts based on the n grams, Section 3 describes the approach of immune systems for clustering, Section 4 shows the experimentation and comparison results and finally Section 5 gives a conclusion and perspectives.

## 2. State of the art

To implement classification methods we should choice a mode of representation of documents (Sebastiani F., 2002). Also, it is necessary to choose a similarity measure and an algorithm for clustering, because there is currently no method of learning that can directly represent unstructured data (text)

**(a) Representation of texts**

A document (text) $d_i$ is represented by a numerical vector as follows:

$$d_i = (V_{1i}, V_{2i}, ..., V_{|T|i})$$

Where T is the set of terms (or descriptors) that appear at least once in the corpus. (|T| is the vocabulary size), and $V_{ki}$ is the weight (or frequency).

- The simplest representation of textual documents is called a "bag of words" representation (Eikvil, L., 1999), it is to transform texts in vectors where each element represents a word. This representation of texts excludes any form of grammatical analysis and any notion of distance between words.

- Another representation, called "bag of phrases", provides a selection of phrases (sequences of words in the texts, not the lexeme "phrases"), by favouring those who are likely to carry a meaning. Logically, such a representation should provide better results than those obtained by the representation "bag of words".

- Another method is based on stemming; it is to seek the root of a lexical term [7] for example, the infinitive forms of the singular for verbs and nouns.

- Another representation, which has several advantages (mainly, this method treats the textual regardless of the language), is based on "n-grams" (a "n-gram is a sequence of n consecutive characters).

There are different methods to compute the weight $V_{ki}$ knowing that for each term, it is possible to compute not only its frequency in the corpus, but also the number of documents containing that term.

Most approaches [9] focus on the vector representation of text using the measure TF∗IDF. TF represents "Term Frequency": the number of occurrences of the term in the corpus. IDF represents the number of documents containing the term. These two concepts are combined (by product) to assign a higher weight to terms that often appear in a document and rarely in the entire corpus.

**(b) Similarity measure**
Several measures of similarity between documents have been proposed in literature in particular is the Euclidean distance, Minkowsky 4 and Cosine that we detail in Section 3.

**(c) Clustering methods**
The principle of clustering is to group texts that seem similar (having common affinities) in the same class. The texts in different classes have different affinities.

The clustering of a set of documents is a highly combinatorial problem. Indeed, the number of possible partitions Pn k, n documents into k classes is given by the Stirling number of second species:

$$P_{n,k} = \frac{1}{k!} \sum_{k} C_k^i (-1)^{k-i} i^n$$

Clustering methods can be divided into two families: the family of methods of hierarchical classification and methods of non-hierarchical classifications.

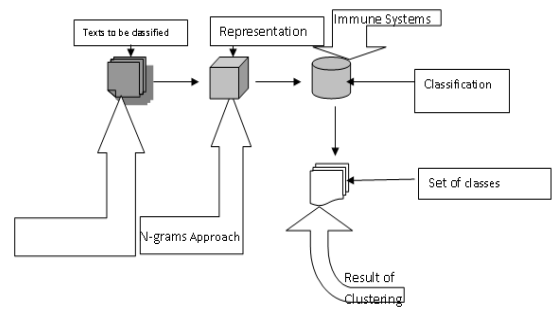

Figure 1: Clustering methods

Unsupervised classification or clustering is a fundamental technique in data mining (structured or unstructured). Several methods have been proposed:
- Hierarchical classification: tree of classes.
- Ascending hierarchical classification: Successive Agglomeration.
- Descending hierarchical classification: Successive divisions.
- Flat Classification: algorithm for k-means: Partition

### 2.1. Biomimetic methods used in classification

In the 70s, the early works on artificial evolution have affected the genetic algorithms (GA), evolution strategies (ES) and evolutionary programming (EP). These three types of algorithms are used broadly common principles because they are all inspired by neo-Darwinism. However methodological choices initially opposed these methods. So the first Genetic Algorithms used rather a binary encoding, but strategies evolution using a real coding. Then in the 90s appeared the genetic programming that introduces particular tree representations. For all these approaches, the representation will also require individual operators to generate new solutions.

#### 2.2.1 Genetic Algorithm
The first work proposing a genetic algorithm to the problem of classification is the work of H. Azzag and al. The number of classes is fixed in advance and the representation of length n associated with each data one class as in the following table1:

| Data  | $d_1$ | $d_2$ | … | $d_n$ |
|-------|-------|-------|---|-------|
| Class | 3     | 1     | … | 4     |

Table 1: The proposed method by H. Azzag and al.

#### 2.2.2 Artificial ants
Lumer and Faieta have proposed an algorithm using a measure of dissimilarity between objects as a distance (Euclidean). The objects correspond to points of a digital space of dimension m, are immersed in a discrete space of smaller dimensions (dimension 2). This discrete space then there is a grid G which each cell can contain an object. Agents move on a region G and receive $R_s$ of s x s boxes in their neighbourhood. The following figure gives an example of a grid with an ant (represented by X) and its detection area (in thick line).

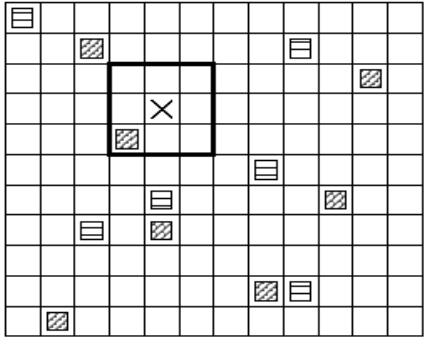

Figure 2 : Grid used in [2]

**Hybridization of the ant colony algorithm and K-means**

An extension of the algorithm Lumer and Faieta was presented by N. Monmarché. On the one hand, ants can stack objects on top of each other in a single grid cell. When they meet a bunch of objects, so they can grab the most dissimilar object. On the other hand, hybridization was performed with the K-means algorithm. This hybridization is to use the following sequence of algorithm: AntClass, K-means, AntClass, K-means. AntClass provides an initial partition, the K-means correcting errors that would AntClass much more time to be corrected with AntClass alone.

**Chemical identification system in ants**

In [14] was introduced a new model based ants for classification, using the system for chemical identification of ants. It is based on the construction of a colonial odour is the result of genetic contributions, environmental and behavioural. This fragrance is built by to identify individuals who belong to the group and must be rejected. From this model, a new classification algorithm has been proposed in which each data is an ant that the smell is determined by the values of the attributes describing this data. The ants perform random encounters and decide to belong to the same group or not. It follows the establishment of a classification.

**Self assembly of ants**

In [11] was introduced a new model to quickly perform a hierarchical classification. It should copy how ants build living structures clinging to each other based on local criteria (the shape of the structure influence the conduct of hanging or dropping out). In this model, each artificial ant represents data. The ants are initially placed at the root of the tree and are able to move around the tree and cling to build a hierarchical structure, each node represents a data. The goal is to automatically build a portal site (text data) and obtain the following property: Each node o of the tree is a class consisting of all data carried by subtrees o. Sub categories (represented by nodes connected to o) should be very dissimilar to their parent tree, but also the most dissimilar between them. The results obtained are very competitive compared to hierarchical classification in particular.

### 2.2.3 Clouds of insects

Principles used for classification, by agent clouds. The agents are initially placed with coordinates and velocity vectors random (see (a)).
The movements of an agent depend on other agents seen in his neighbourhood and similarities between the data they represent. The local behaviour of each agent tends to form aggregate groups of similar agents moving in a consistent manner (see (b)).

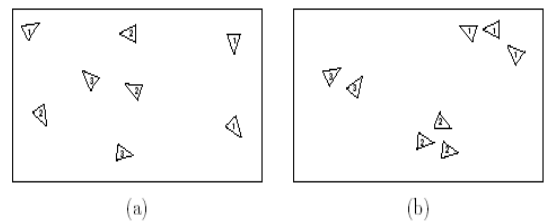

Figure 3 : Principle clouds of insects

### 2.2.4 Swarm intelligence

Swarm intelligence includes many algorithms based population of agents. The algorithms that interest us are those that use displacements of a swarm of agents to solve a problem. For example, the algorithms PSO (particle swarm optimization) using a set of particles characterized by their position and speed to maximize a function in space research. Interactions occur between particles in order to achieve effective global behaviours.
In biology, many researchers are interested in how animals move in groups. No one individual controls the other but forms and complex behaviours can occur during these trips.
[24] was probably the first to propose a use of such computer models, simulations are mainly used in the film industry to give realistic movement to groups of individuals. In this work, each individual evolves in 3D space. It is characterized by its position and speed. An individual perceives the other in a given neighbourhood. Simple behavioural rules allow individuals to travel in groups, avoid obstacles, etc.

In 1998, these principles have been applied for the first time in classification problems (see Figure 3). Agents representing each data. An agent responds to other agents in its neighbourhood, taking into account the similarity of data. An agent will move instead to data that are similar. This behavioural rule can then form groups of similar data. In [11], this algorithm has been improved and systematically evaluated. An ideal distance between individuals is defined distance depends on the similarity between data. A stopping criterion is also used in measuring the entropy of the agent cloud. This algorithm has been integrated into a system of visual data mining using virtual reality.

**2.2.5 Cellular Automata**

In [19] the cellular automaton is a grid cell structure with a flat neighbourhood arising from this structure (planar). Three functions of transitions were used to vary the automaton with four states for each cell. The results obtained show that the virtual machine parallel computing (Class_AC) effectively includes similar documents on near threshold.

## 3. Representation of texts based on n-grams

An n-gram is a sequence of n consecutive characters in a document, all n-grams (usually n = (2, 3, 4.5)) is the result obtained by moving a window of n boxes on the body text. This movement is made by steps of one character and each step we take a picture. All of these pictures give the set of all n-grams of the document. Then there are the frequencies of n-grams found.

For example, to generate all 5-grams in the sentence "The girl eats the apple" we get:

The-g, he-gi, e-gir, -girl, girl-,irl-e, rl-ea, l-eat, -eats, eats-, ats-t, ts-th, s-the, -the-, the-a, he-ap, e-app, -appl, apple.

Note
. A dash (-) represents the space between words in text

Since its creation by (Shannon) in 1948, the concept of n-grams has been widely used in several areas such as the identification of speech and information retrieval: representation of textual document by the method of n-grams has many advantages In fact, the n-grams capture the knowledge of most frequent words of each language which facilitates identification of language and the method of n-grams is independent of language, while the systems based on words For example, are dependent on language.

Another advantage of the representation of texts with the n-grams is that this method is tolerant to spelling mistakes, for example, when a document is scanned using an OCR, the OCR is often imperfect.

For example, the word "character" can be read as "claracter. A system based on the words hardly recognize the word "character" or even its root; But, a system based on the n-grams will be able to take into account other n-grams as " aract", "racte" etc., with n = 5, some retrieval systems based on n-grams have retained their performance with the deformation rate of 30%, a rate at which any system based on words cannot function properly. Finally, with the use of n-grams for the representation of textual documents is not required to pre-treatment language, that is to say, the application of techniques stemming or elimination of "stop words", sequences of n-grams do not improve performance.

For example, if a document contains many words from the same root, the frequencies of n-grams corresponding increase without requiring any prior language processing. In the phrase "the fisherman fishing" the 5-grams and the corresponding representative vector are as follows:

| N-gram | Number of occurrence |
|--------|----------------------|
| The-f  | 1 |
| he-fi  | 1 |
| e-fis  | 1 |
| -fish  | 2 |
| Fishe  | 1 |
| Isher  | 1 |
| Sherm  | 1 |
| Herma  | 1 |
| Erman  | 1 |
| Rman-  | 1 |
| Man-f  | 1 |
| an-fi  | 1 |
| n-fis  | 1 |
| -fish  | 2 |
| Fishi  | 1 |
| Ishin  | 1 |
| shing  | 1 |

Table 2: The 5-grams corresponding to the phrase "the fisherman fishing" with their representative vector.

### 3.1 Advantages of techniques based on n-grams

- Compared with other techniques, the n-grams automatically capture roots of most frequent words. One need not step Research lexical roots.

- They operate independently of language, unlike systems based on words in which it must use specific dictionaries (feminine masculine, a singular, plural, conjugation…) for each language.

- With n-grams, we do not need prior segmentation of text into words; this is interesting for the treatment of languages in which the boundaries between words are

- not strongly marked, such as Chinese, or for DNA sequences in genetics.
- They are tolerant to spelling mistakes and distortions caused during the use of scanners. When a document is scanned, the OCR is often imperfect. For example, it is possible that the word "chapter" is read as "clapter. A system based on the words will have difficulty recognizing it is the word "chapter" because the word is misspelled.
- Finally, these techniques do not need to remove function words (Stop Words) or make lemmatization (Stemming). These treatments increase the performance systems based on the words. By cons, for the systems n-grams, many studies [7] showed that the performance does not improve after the elimination the "Stop Words" and "Stemming".

### 3.2 Dimensionality reduction

Most learning algorithms are not tolerant to the large size of the space of representation. It is a very common problem in the field of automatic classification of texts, for example, with the representation of texts in bag of words, each word of the corpus is a potential descriptor, so a corpus of reasonable size, this number may be several tens of thousands, the idea of selecting a subset of these descriptors is therefore essential.

The deletion of certain words, such as words that appear infrequently in the corpus, because their low frequencies do not provide information, not resolve the problem of the large size of the representation space, so even after removing certain words the number of descriptors is high, the solution is to use a statistical method for selecting relevant words which helps to easily discriminate between classes of documents.

The techniques used for dimension reduction are derived from information theory and linear algebra; we can distinguish, generally, two types of these techniques; the first by selecting terms and second by extracting terms. [25]

The method used in our study is that of $\chi^2$.

A statistical measure known, it adapts well to the selection of attributes because it evaluates the lack of independence between a word and a class. It follows the following approach:

- Be the pivot table, $N_{ij}$, occurrences of n-gram i in text j;
- Calculate the frequencies f ij corresponding by $f_{ij} = N_{ij} / N$;
- Calculate the contributions of (ij) to the statistics of $\chi^2$:

$$X_{ij}^2 = \frac{\left(N_{ij} - \frac{N_i * N_j}{N}\right)^2}{\frac{N_i * N_j}{N}}$$

- Sort the table of $\chi^2$ in order uncrossing;
- Finally, determine the list (gram i) of the first k (n-grams) for each text for standardized.

After reduction of the size we have obtained for each gram a reduction rate summarized in the following table:

| Number of Document :400 ||||
|---|---|---|---|
| N-Grams | Number of term before Reduction | Number of term after Reduction | Rate Reduction |
| 2 | 528 | 432 | 18,18% |
| 3 | 3799 | 1045 | 72,49% |
| 4 | 9461 | 1105 | 88,32% |
| 5 | 11707 | 1055 | 90,98% |

Table 3: reduction rate

### 3.3 Digitalization

The digitalization is realised by the method TF-IDF (Term Frequency / Inverse Document Frequency) which is derived from an algorithm for information retrieval. The basic idea is to represent documents by vectors and measure the closeness between documents by the angle between vectors, this angle is assumed to represent a semantics distance. The idea is to encode each word of the bag by a scalar (number) called tf-idf to give a mathematical aspect to text documents.

$$tfidf = tf(i,j).idf(i) = tf(i,j).\log\left(\frac{N}{N_i}\right)$$

Where:
- tf(i,j) is the term frequency: the frequency of term ti in document dj
- idf(i) is the inverse document frequency: the logarithm of the ratio between the number N of documents in the corpus and Ni the number of documents containing the term ti.

A document corpus di after digitalization is:
di = (x1, x2, ... ...., xm) where m is the number of word of ith bag of word and xj is the tf-idf.

This indexing scheme gives more weight to words that appear with high frequency in some documents. The underlying idea is that these words help to discriminate between texts with different subject. The tf-idf has two fundamental limitations: The first is that longer documents are typically rather strong weight because they contain more words, so "the term frequencies" tend to be higher. The second is that the dependence of the "term frequency" is too important. If a word appears twice in a document dj, it does not necessarily mean that it has two times more important than in a document dk where it appears only once.

In our study we standardized frequencies TFxIDF by the following formula:

$$TF \times IDF\ (t_k, d) = \frac{n_i}{\sum_i n_i} * log\left(\frac{N}{DF}\right)$$

The weighting of TF × IDF has the following effects:

1. Importance of each word in the standard text
2. A word that appears in all documents is not important for differentiation of texts
3. Relevance words globally uncommon but common in certain documents. Encoding TF × IDF does not correct the lengths of texts, to this end, the coding TFC is similar to that of TF × IDF, but it fixes the lengths of the texts by cosine normalization, in order not to encourage the longest.

$$TFC\ (t_k, d) = \frac{TF \times IDF\ (t_k, d)}{\sqrt{\sum_{s=1}^{|r|} (TF \times IDF\ (t_s, d))^2}}$$

## 4. The immune systems for clustering

### 4.1 The similarity matrix

We experiment the classification using three type of similarity distance: the Euclidean distance, Minkowsky 4 distance and cosine distance. For each distance we construct a triangular symmetric similarity matrix where each element S (i, j) corresponds to the distance between document i and document j and S (i, i) = 0 for Euclidean and monkowsky 4 distances and s ( i, i) = 1 for the cosine distance.

**Euclidian distance or Minkowsky 2**
Distances between vectors $T_i$ and $T_j$ in multidimensional space are:

$$D(T_i, T_j) = \sqrt{\sum_k (x_k(T_i) - x_k(T_j))^2}$$

**Minkowsky 4 distance**
Distances between vectors $T_i$ and $T_j$ in multidimensional space are:

$$D(T_i, T_j) = \sqrt[4]{\sum_k (x_k(T_i) - x_k(T_j))^4}$$

**Cosine distance**
Distances between vectors $T_i$ and $T_j$ in multidimensional space are:

$$Cos(T_i, T_j) = \frac{T_i . T_j}{||T_i|| . ||T_j||}$$

Where $T_i$, $T_j$ represents the scalar product of the vectors $T_i$ and $T_j$
$||T_i||$ and $||T_j||$ Represent respectively the standards of $T_i$ and $T_j$.

The similarity matrix is a symmetric matrix of dimension N ∗ N, where N is the number of documents to classify diagonal equal to zero for the Euclidean distance and Manhattan and diagonal equal to 1 for the cosine distance. The indices represent the indexes in the corpus of the documents to classify.

### 4.2 Immune System

The term "immune system artificial" (Artificial Immune System, AIS) applies to a wide range of different systems, including meta-heuristic optimization inspired by the functioning of the immune system of vertebrates. Many systems have been designed in several different areas such as robotics, detection of anomalies or optimization. The immune system is responsible for protecting the body against "aggressions" of external bodies. The metaphor from which the AIS algorithms focuses on aspects of learning and memory of the adaptive immune system says (as opposed to so-called innate), particularly through the discrimination between self and no self. Indeed, living cells possess on their membranes of specific molecules called "Antigens". Each body therefore has a unique identity, determined by all the antigens present on cells. The lymphocytes (a type of white blood cells) are immune cells that have receptors able to bind specifically to a single antigen, thereby to recognize a foreign cell in the body. A lymphocyte recognized as having a cell non-self will be stimulated to proliferate (and producing clones of himself) and differentiate themselves in their cells to keep in memory the antigen, or cell to combat the attacks. In the first case, it will be able to react more quickly to a new exhibition of antigen: it is the very principle of efficacy of vaccines. In the second case, the fight against the attacks is possible by the production of antibodies.

Figure 4 summarizes the main steps. It should also be noted that the diversity of receptors in the whole lymphocyte population in turn is produced by a mechanism of hyper-mutation of cloned cells.

Immune systems (IS) are a set of models of human and animal immune system applied to different computing problems. They use the following principles: agents (lymphocytes) that produce antibodies will learn to recognize self from non-self (antigens). For this, these agents must first be generated using a composition principle of building blocks. Then, they undergo a selection test (called negative selection): agents rejecting the self are eliminated, and others who will reject non-self, are kept. Each time there is recognition of an antigen by an antibody, the presence of lymphocytes that generate these antibodies is favoured by a selection process by cloning and by the disappearance of lymphocytes not stimulated by antigens. This cloning gives rise to interactions between lymphocytes and can implement mutations. Some lymphocytes, when used often, take a role of storage element in the long term. These systems have complex properties because they are able to generate solutions and select them according to their effectiveness as original heuristics.

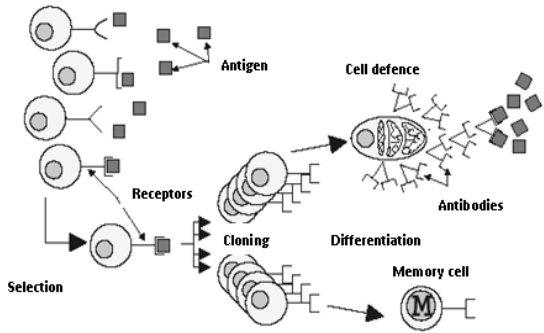

Figure 4: The selection by cloning: lymphocytes presenting specific receptors for antigen are differentiated into memory cells or cells participating in the active defence of the body by means of antibodies.

### 4.3 Approach for clustering

Data: $d_1,...,d_n$ represent antigens. These antigens are presented to the system repeatedly until a stop condition. The modelling data are digital, and therefore that the antigen is a vector of dimension n. At each iteration, the antigen is presented to enable antibodies (similar, in this modelling, lymphocyte -B). An antibody is also represented by a vector of dimension n. The antibodies close enough to the antigen (in the sense of Euclidean distance) (ie: similar) will undergo clones with mutation (interaction antibody / antigen) to amplify and refine the system response. Also, these antibodies will undergo a selection (interaction antibody / antibodies): those who are too close to each other will be reduced in number. After these iterations, the system converges by placing antibodies (which act as sensors) wisely and number of suitable data.

The main ideas used for the design of metaheuristics are operated upon the selections on lymphocytes, accompanied by the positive feedback allowing the multiplication and the memory system. Indeed, these attributes are critical to maintain the properties of self-organized system

### 4.4 Algorithm

We retrieved 400 documents of the Reuters 21578 corpus (20 documents of each file extension ".sgm") which was digitized by the TF-IDF method. T1 to t400 texts represent antigens (an antigen is a vector of dimension k where k is the number of terms because the method of indexation used is bag of words). An antibody is also represented by a vector of dimension k.

- Indexing documents corpus to classify (N-Grams Approach).
- Digitalize each text document corpus by the TF-IDF method (Digitalization).
- Compute the similarity matrix from the vectors found: sim (i, j) = D (di, dj) where i,j .are document vector and D is distance used.

  While the stop condition is not verified do
    For i= 1 to n do begin
If antibody is similar to antigen (D (i,j) ≅ 0 or D (i,j) ≅ 1)
"The distance is compared to 1 in the case of the cosine distance"
"The distance is compared to 0 in the case of the Euclidian and Minkowsky4 distance"
Then begin
        To cloning antibody and Do Mutation "Interaction antibody / antigen"
        Do selection "Interaction antibody / antibody"
        Antibodies that are too close to each other will be reduced in number
                End If
                End For
Reconstruction of the similarity matrix with the new antibodies found
        End While

### 5. Experimentations

After testing the algorithm on the document of the Reuters 21578 corpus, we obtained the following results in terms of number of classes and purity of the clusters:

In terms of purity of the cluster, and error rate in classification we used one measure of assessment in this case the f-measure that is based on two concepts: recall and precision defined as follows:

$$precision(i,k) = \frac{N_{i,k}}{N_k}$$
$$recall(i,k) = \frac{N_{i,k}}{N_{c_i}}$$

where N is the total number of documents, i is the number of classes (predefined), K is the number of clusters, NCi is the number of documents of class i, Nk is the number of documents to cluster Ck , Nik is the number of documents of class i in cluster Ck. The f-measure is calculated on a partition P as follows:

$$F(p) = \sum \frac{N_{C_i}}{N} \max_{k=1}^{K} \frac{(1+\beta) \times recall(i,k) \times precision(i,k)}{\beta \times recall(i,k) + precision(i,k)}$$

In general β = 1

The partition P corresponds to the expected solution is one that maximizes the F-measure or minimizes the associated entropy. (In our study P is the partition that corresponds to the class of results of classification by the cellular automata method for the number of documents associated).

### 5.1 Results

We experienced our immune systems on the Reuters 21578 corpus, we proceeded to the extraction of 400 texts we have indexed and then calculated the similarity matrix. After testing we have achieved the results grouped in table 2.

### 5.2 Interpretation

The experiment started with 50 documents and the results have only proved that immune systems were able to make the clustering of text documents (Clustering of text) because they regrouped effectively similar documents. Then we performed the experiment with 150 documents (first 50 texts of three documents REUTERS 21,578) and we have achieved concrete results from the previous, which led us to increase the size of the corpus documents in 400 to take a decision on the quality of the classification after evaluation by the F-measure.

The colored boxes in the tables above represent the best classification. The F-measure are highlighting in blue. Regarding the choice of the best classification we opted for the F-measure because it is based on two concepts (recall and precision) as shown in formula (1).

In terms of time, the convergence of the algorithm depends on the number of gram. The number of gram increase over the convergence time increases.

### 5.3 Comparison

Our results were compared with an clustering performed by cellular automata for the same database REUTERS 21,578 and the same number of documents (400 documents) [20] (see Table 3).

Based on the values of the f-measure, the method of immune systems has an influence, because the results of the evaluation are much higher. But based on the time of the clustering, the method of cellular automata is faster than the immune systems.

As regards distances and based on the f-measure, cosine distance gives the best results (highlighted in blue) but in terms of number of class and then by manual expertise, the Euclidean distance to the immune systems is the most appropriate (highlighted in yellow).

| **Distance** | **Cosine** | | | | **Euclidian** | | | | **Minkowsky 4** | | | |
|---|---|---|---|---|---|---|---|---|---|---|---|---|
| **Grams** | 2 | 3 | 4 | 5 | 2 | 3 | 4 | 5 | 2 | 3 | 4 | 5 |
| **# Class** | 2 | 25 | 35 | 38 | 6 | 31 | 30 | 41 | 38 | 57 | 46 | 52 |
| **Clustering Time(ms)** | 2075 | 2089 | 7977 | 9452 | 2063 | 2400 | 8125 | 9112 | 1545 | 4595 | 9125 | 10490 |
| **F-measure %** | 49,0 | 35,16 | 41,12 | 34,0 | 40,5 | 33,01 | 45,33 | 23,33 | 33,12 | 31,74 | 23,0 | 20,05 |

Table 4: Result of classification (Immune System)

| | **Cellular Automaton** | | | **Immune systems** | | |
|---|---|---|---|---|---|---|
| | Cosine | Euclidian | Manhattan | Cosine | Euclidian | Monkowsky4 |
| Number of Class | 157 | 42 | 42 | 2 | 30 | 38 |
| F-measure | 40% | 27% | 17% | 49% | 45,33% | 33,12% |
| Learning Time (s) | 0,386 | 0,243 | 0,342 | 2,075 | 8,125 | 1,545 |

Table 5: Comparison with cellular automata

Note: For comparison and for different distances we have taken the best results regardless of grams used. We are based only on the evaluation criterion (F-measure).

## 6. Conclusion and Perspectives

We proposed a first algorithm for clustering using immune systems. After testing we have proved that this algorithm can solve a text mining problem .i.e. the clustering.

The methods of indexing text documents such as TF-IDF and n-grams approach helped us to digitalizes documents so that the use of immune systems on digital vectors. So passage of documents to text, text to number, number to the analysis by immune systems and analysis to decision making on the classification have been the subject of this study. This algorithm was compared with an algorithm using learning by

KOHONEN self organizing maps. Our approach can contribute in solving data mining problem (clustering). As future work; we are studying immune systems with other parameters. The algorithm will also be tested for other types of data such as images and multimedia data in general.